\documentclass[twoside]{article} 

\usepackage{hyperref}
\usepackage{amsfonts}       %
\usepackage{nicefrac}       %
\usepackage{graphicx}
\graphicspath{{fig/}}
\usepackage{caption}
\usepackage{subcaption}
\usepackage{rotating}

\usepackage{multirow}

\usepackage{delarray,mathtools}

\usepackage{natbib}

\usepackage{algorithm}
\usepackage{algpseudocode}
\usepackage{fullpage}

\usepackage{amsmath, amssymb, amsthm, tikz}

\usepackage[utf8]{inputenc} %
\usepackage[T1]{fontenc}    %
\usepackage{hyperref}       %
\usepackage{url}            %
\usepackage{booktabs}       %
\usepackage{amsfonts}       %
\usepackage{nicefrac}       %
\usepackage{microtype}      %

\usepackage[titletoc, title]{appendix}
\usepackage{graphicx}
\graphicspath{{fig/}}
\usepackage{caption}
\usepackage{subcaption}
\usepackage{rotating}

\usepackage{multirow}

\usepackage{delarray,mathtools}

\usepackage{natbib}

\usepackage{bm}

\usepackage{algorithm}
\usepackage{algpseudocode}

\usepackage{hyperref}

\usepackage{amsmath, amssymb, amsthm, tikz}
\usepackage{blkarray}

\newcommand{\cO}{\mathcal{O}}

\newcommand{\bR}{\mathbb{R}}

\newcommand{\diag}{\mathrm{diag}}

\newcommand{\df}{\overset{\text{def}}{=}}

\begin{document}

\title{Rapid Posterior Exploration in \\ Bayesian Non-negative Matrix Factorization}

\author{M. Arjumand Masood and Finale Doshi-Velez}

\maketitle 

\begin{abstract}

Non-negative Matrix Factorization (NMF) is a popular tool for data
exploration. Bayesian NMF promises to also characterize uncertainty in
the factorization.  Unfortunately, current inference approaches such
as MCMC mix slowly and tend to get stuck on single modes.  We
introduce a novel approach using rapidly-exploring random trees (RRTs)
to asymptotically cover regions of high posterior density. These are
placed in a principled Bayesian framework via an online extension to
nonparametric variational inference. On experiments on real and
synthetic data, we obtain greater coverage of the posterior and higher
ELBO values than standard NMF inference approaches.
\end{abstract}

\section{Introduction}
Non-negative Matrix Factorization (NMF) is a popular model for
understanding structure in data, with applications such as
understanding protein-protein interactions \cite{Greene}, topic
modeling \cite{roberts2016navigating}, and discovering molecular
pathways from genomic samples \citep{brunet2004metagenes}.  The goal
is simple: given a $D \times N$ data matrix $X$ and desired rank $R$,
the nonnegative matrix factorization (NMF) problem involves finding an
$R \times N$ nonnegative weight matrix $W$, and an $D \times R$
nonnegative basis matrix $A$, such that $X \approx AW$.  Applied work
has benefitted from the myriad of efficient algorithms for solving
different versions of the NMF objective
\citep{paisley2015bayesian,schmidt2009bayesian,moussaoui2006separation,lin2007projected,lee2001algorithms,recht2012factoring}.

However, in many cases NMF is not identifiable: there may be different
pairs $(A,W)$ and $(A',W')$ that might explain the data $X$ (almost)
as well.  Non-identifiability in NMF has been studied in detail in the
theoretical literature
\citep{pan2016characterization,donoho2003does,arora2012computing,intersectingfaces15,bhattacharyya2016non},
but it is of practical concern as well.  For example, \cite{Greene}
use ensembles of NMF solutions to model chemical interactions, while
\cite{roberts2016navigating} conduct a detailed empirical study of
multiple optima in the context of extracting topics from large
corpora.  Both works use random restarts to find multiple optima.

Bayesian approaches to NMF \citep{schmidt2009bayesian,
  moussaoui2006separation} promise to characterize parameter
uncertainty in a principled manner by solving for the posterior
$p(A,W|X)$ given priors $p(A)$ and $p(W)$.  Having such a
representation of uncertainty in the bases and weights can further
assist with the proper interpretation of the factors: we may place
more confidence in subspace directions with low uncertainty, while
subspace directions with more uncertainty may require further
exploration.  Unfortunately, in practice, these uncertainty estimates
are often of limited use: variational approaches
(e.g. \citet{paisley2015bayesian, hoffman2015structured}) typically
underestimate uncertainty and fit to a single mode; sampling-based
approaches (e.g. \citet{schmidt2009bayesian, moussaoui2006separation})
also rarely switch between multiple modes.

In this work, we take steps to a more complete characterization of
uncertainty in NMF.  First, we use recent insights
\citep{pan2016characterization} about how different NMF solutions of
similar quality may be related to each other to limit the spaces that
one must explore to find alternate solutions.  Second, we use
rapidly-exploring random trees (RRTs) with an online nonparametric
variational inference framework to rapidly cover the space of probable
solutions.  The first part is specific to NMF; the second is broadly
applicable to other statistical models as well.  We demonstrate that
our approach achieves not only better posterior coverage (as measured
by the covering number metric of
\citet{DBLP:journals/corr/MasoodPD16}), but often better ELBO values
as well.

\section{Background}

\paragraph{Nonparametric Variational Inference (NVI)}
Variational methods approximate a desired distribution $p(\theta|x)$
with a distribution $q(\theta)$ by maximizing the evidence lower bound
(equivalent to minimizing $\texttt{KL}(q||p)$):
\begin{equation}
\mathcal{F}[q] = \mathcal{H}[q] + \mathbb{E}_q[\log p(\theta,x)]. 
\label{eqn:elbo} 
\end{equation} 
To model complex patterns in $p(\theta|x)$,
\citet{gershman2012nonparametric} suggest using a mixture of uniformly
weighted $M$ gaussians with isotropic co-variance for the variational
family $q(\theta)$: 
\begin{equation*}
q(\theta) = \frac{1}{M} \sum_{m = 1}^{M} \mathcal{N}(\theta; \mu_m, \sigma_m^2\bm{I})
\end{equation*} 
The entropy term $\mathcal{H}[q]$ encourages diversity in the set of components. The expectation term $\mathbb{E}_q[\log p(\theta,x)]$ encourages the components to be placed in regions of high joint likelihood. This can be viewed as a quality component of the ELBO. 
The authors note that the entropy term $\mathcal{H}[q]$ in equation~\ref{eqn:elbo}
can be lower bounded by
\begin{equation} 
\label{eqn:ent}
\mathcal{H}[q] \geq -\frac{1}{M} \sum_{m = 1}^{M} \log \left(  \frac{1}{M} \sum_{j= 1}^{M} \mathcal{N}(\mu_m ; \mu_j, (\sigma_m^2 + \sigma_j^2)\bm{I}) \right)\end{equation}  
and provide a second-order approximation for the likelihood term
$\mathbb{E}_q[\log p(\theta,x)]$, where we use $f(\theta) = \log p(\theta,x)$:

\begin{equation}
\label{eqn:2ndorder} 
\mathbb{E}_q[f(\theta)] \approx \frac{1}{M} \sum_{m = 1}^{M} f(\mu_m) + \frac{\sigma_m^2}{2}\text{Tr}(\bm{H}_m)
\end{equation} 
where $\bm{H}_m$ is the Hessian $\nabla^2_\theta f(\theta)$.  While
many other approaches could be used to optimize this term
(e.g. BBVI~\cite{ranganath2014black}), we will use the approximation
from equation~\ref{eqn:2ndorder} as it permits certain aspects of our
optimization to be computed analytically.

\paragraph{Rapidly-Exploring Random Trees}
Rapidly-Exploring Random Trees (RRTs) and their variants have been
extremely successful for solving high-dimensional path-planning
problems in robotics and other domains \cite{lavalle1998rapidly}.
RRTs take as input a configuration space $\mathfrak{C}$---for example,
all possible parameter settings---which may have some obstacles
$\mathfrak{O} \subset \mathfrak{C}$ or invalid parameter settings.
Given a metric $\rho(c,c')$ on the configuration space $\mathfrak{C}$
, a method for uniformly sampling within $\mathfrak{C}$, a means to
determine if a specific point $c \in \mathfrak{O}$, and a current set
of nodes $\{c\}$ is in the tree, the RRT rapidly expands across the free
space via the following algorithm:
\begin{enumerate}
\itemsep0em
\item Generate a sample $c^* \in \mathfrak{C}$ uniformly 
\item Find the node $c \in \{c\}$ that is nearest to $c^*$ according
  to the metric $\rho(c,c^*)$.
\item Add the node $c'$ that is obtained by moving a step-size
  $\epsilon$ from $c$ in the direction of $c^*$ if $c' \notin
  \mathfrak{O}$.
\end{enumerate}
This procedure implicitly expands nodes in proportion to the volume of
the Voronoi region associated with each node in the configuration
space $\mathfrak{C}$, making the RRT seek new regions.  In the context
of NMF posterior characterization, we will use geometric insights to
first reduce the size of the configuration space to be explored, and
then apply RRTs to rapidly explore this reduced space.  The resulting
samples will be used as centers for a NVI approximation of the posterior.

\paragraph{Covering Number as a Metric for Posterior Coverage}
Even for relatively small amounts of data (e.g. 100 observations),
sampling-based procedures for Bayesian NMF can be mixed within a
mode---as measured by traditional mixing metrics such as
autocorrelations times---but slow to explore even a single mode.  What
is missing is a notion of coverage: In addition to moving in
``independent" ways, how much of the posterior space does a
finite-length chain explore?  \citet{DBLP:journals/corr/MasoodPD16}
advocate for the use of the minimum covering number to quantify how
much of the posterior space a finite-length chain explores. Given a
similarity measure on the parameter space, the \emph{covering number}
is the minimum number of $\epsilon$ balls needed to cover all the
samples.  We approximate the covering number by limiting our centers
to existing samples and greedily adding centers until all the points
are covered.  Larger covering numbers suggest larger exploration of
the posterior space.  Finally, a \emph{persistance plot} shows how the
covering number changes with $\epsilon$; for points that are spread
far apart, the covering number will remain large for many $\epsilon$.

For the similarity measure, we use weighted angular distance (WAD;
an extension of the maximum angle similarity proposed in \citet{DBLP:journals/corr/MasoodPD16}).  To compute
the WAD, we first compute an angle difference between columns
of the basis matrix after adjusting for the permutation ambiguity in
NMF.  Next we consider the contribution of each vector in the cone in
reconstructing the data (weights). Given two pairs of factorizations,
$(A, W)$ and $(A', W')$, let $\widehat{\rho}$ be a permutation of the
columns of $A'$ that minimizes the average angle between corresponding
columns,
\begin{align*}
\widehat{\rho} = \arg\min_{\rho\in S_R} \frac{1}{R}\sum_{r \in R}
\cos^{-1}\left(\frac{A_r \cdot
  A'_{\rho(r)}}{\|A_r\|\|A'_{\rho(r)}\|}\right)
\end{align*}
Let $\bm{\alpha}$ be the the vector of angle differences between
corresponding columns of $A$ and $A'$. In order for the measurement to
be consistent across different scalings of the factorizations, we fix
the basis matrices to have the scaling that makes them column
stochastic and now consider the corresponding weights matrices $W$ and
$W'$. We sum the rows of the two weights matrices and normalize them
to add to 1. We denote these weights vectors as $\bm{w}$ and
$\bm{w'}$. The WAD is then given by
\begin{align*}
\text{WAD}(A,W,A',W') = \bm{\alpha}^T\frac{(\bm{w} + \bm{w'})}{2}
\end{align*}
The WAD is bounded in the interval $[0, 90]$ degrees since all the
columns of $A$ and $A'$ lie in the positive orthant.  WAD is also
invariant to the scaling and permutation ambiguity in NMF.

\section{Methods}
We now detail our approach, which leverages geometric insights from
the NMF problem to use RRTs to rapidly explore nodes to be
incorporated into an online NVI framework.

\subsection{Online Nonparametric Variational Inference}
This work is built upon the work in
\citep{gershman2012nonparametric}. We use the a more general
variational family by allowing weights $w_m$ to be non-uniform.  We
also modify the algorithm to an on-line one which can be used to
decide whether a new mixture component should be included in
the variational distribution or not. Our Online-NVI (ONVI) takes in a
candidate component center $\mu_{m+1}$ and does the following:
\begin{enumerate}
\item Find the optimal member of the variational family with the added
  component (optimizing for $\sigma^2_{m+1}$, $w_{m+1}$).
\item Decide whether to add the new component.
\item If we accept a new component, decide whether to remove previous
  components.
\end{enumerate}
In particular, for the first step, we assume that the previous centers
$\mu_1,\dots,\mu_{m}$ and variances $\sigma^2_1,\dots,\sigma^2_{m}$
stay fixed, and that previous weights $w_1,\dots,w_{m}$ are simply
scaled with the addition of a new weight $w_{m+1}$.  This restriction
is reasonable because of mode-hugging property of variational
inference: each component will want to find a mode and will tend to
under-estimate the variance around that mode; thus, it is reasonable
to expect that adding a new component will not cause global changes to
the NVI solution.

For the second and third steps, we note that adding new components can
only make the ELBO increase as we have just created a more flexible
variational family---if the proposed candidate mean $\mu_{m+1}$ is of
low quality, we will simply set its weight $w_{m+1}$ to zero.
Similarly, if a new component is added with significantly higher
quality than the existing components, then it is possible that all
previous weights $w_1,\dots,w_m$ will be scaled down to close to zero.
For both cases, we use a minimum change criterion: a minimum ELBO
improvement criterion to accept a new component, and a minimum ELBO
loss to reject an existing component.  

\subsection{NMF exploration using RRTs}

\paragraph{Framework} While there exist many RRT variations, we use the following framework: 
\begin{enumerate}
\item Base Nodes: A set of nodes that always remain part of the RRT.
\item Temporary Nodes: A set of nodes temporarily used in the
  expansion of the RRT.  They are deleted once a certain expansion
  criteria is met to avoid slowing down future nearest neighbor
  searches.
\item Feasibility Test: A way of checking if a proposed node is in our
  configuration space (that is, our region of interest).  
\item Stepping Method: A way to move from one node in the direction of
  another.  
\end{enumerate}

\paragraph{Configuration Space}  The space of all the parameters needed
for $A$ and $W$ is large---of dimension $R(D + N)$.  In this section,
we use the geometry of the NMF problem to limit the space in which our
RRT will have to search. Let $A_{\text{SVD}} \in \mathbb{R}^{D
  \times R}$, $W_{\text{SVD}} \in \mathbb{R}^{R \times N}$ be the SVD
of $X$.  The SVD factorization is exact when the data $X$ has rank $R$. We note that when the rank $R$ of
a matrix is equal to its positive rank $R^+$, \citet{Laurberg} show
that two exact factorizations can only differ by a subspace
transformation $Q$ of the form
\begin{equation*} 
X = AW = (AQ)(Q^{-1}W) = A_{\text{SVD}}W_{\text{SVD}}  
\end{equation*} 
where the last equality must be true because $A_{\text{SVD}}$ defines
the unique subspace in which the data live.  That is, the NMF problem
can be viewed as a change of basis from the subspace spanned by the
SVD factors.  While the unique-subspace assumption is not necessarily
true for the approximate case, it does give us a way to rapidly
explore the best subspace of rank $R$.  $Q$ has only $R^2$ parameters
(we note that \citep{rovckova2015fast} use a similar insight within an
auxiliary variable approach to find sparse matrix factorizations).  

Two concerns remain: First, the space of $Q$, while smaller, still
contains many trivial transformations: permutations and changes of
scale.  Second, the change of basis $A_{\text{SVD}}Q$ and new weights
$Q^{-1}W_{\text{SVD}}$ may not produce nonnegative factors.  To
address the first concern, we limit our change of basis matrices $Q$
to the oblique manifold which reduces the redundancies from infinite
(permutations and scalings) to a finite set of permutations.  A
rank-$R$ oblique manifold is defined as
\begin{equation*}
\mathcal{O}(R) \df \{ Q \in \mathbb{R}^{R\times R} : \diag(Q^\top Q) = I_R, \text{det}(Q) \neq 0\}
\end{equation*}
 In words, $\cO(R)$ is the space of invertible $R \times R$ matrices
with columns that are unit Euclidean norm.  In fact, the oblique
manifold, $\cO(R)$ can be treated as the product of $R$ unit-spheres
in $\bR^R$: the only difference being that these spheres should all be
linearly independent in the Oblique manifold.  The only remaining NMF
inherent ambiguity is that of the permutation of columns but that is
only finite. For a rank $R$ problem, given a $Q$, we have $R!$
equivalent factorizations that can be obtained through considering
different permutations of the columns of the matrix $Q$---a measure
zero set of points.  

Next, we address the fact that not all matrices $Q$ will produce a
nonnegative solution $A_{\text{SVD}}Q$ and $Q^{-1}W_{\text{SVD}}$.  We
project the products to positive values.  The $\lfloor
\rfloor$ operation stands for setting negative values to zero:
\begin{equation*} 
A = \lfloor{A_{\text{SVD}}Q}\rfloor \qquad 
W = \lfloor Q^{-1}W_{\text{SVD}}\rfloor
\end{equation*} 
(Note that in the case of exact NMF, there exists a $Q$ for which $A =
A_{\text{SVD}}Q$ and $W = Q^{-1}W_{\text{SVD}}$, where there are no
negative values in the products so $A,W$.)  

\paragraph{Initial Base Nodes}  We initialize our RRT with a set of base nodes $Q_m$ corresponding
to each random restart $(A_m,W_m)$ as the change of basis from the
SVD: $Q_m =\text{argmin}_{Q} || A_m - A_{\text{SVD}}Q ||$.  As
the tree is expanded, we only retain nodes that are sufficiently far
from existing nodes; these criteria are detailed in
Section~\ref{sec:imp_details}.

\paragraph{Feasibile Regions} 
We compute feasible regions by thresholding based on the joint
probability of data and factorizations.  We also adaptively adjust the
feasibility criteria via a minimum angle threshold in order to avoid
finding new points that are too close to existing ones.

\paragraph{Stepping Method} 
Movement from one node $Q_1$ to another $Q_2$ is made by stepping in
the direction $Q_d = Q_2 - Q_1$.  Each step consists of moving
along the tangent space and then retracting the result onto the
oblique manifold (e.g. as in \citep{absil2009optimization}).
Following the RRT-Extend algorithm of \citet{kuffner2000rrt}, we
continue until the feasibility criteria fails or a maximum number of extended steps are tried. 

\paragraph{Aside: Scale Optimization for Gaussian Likelihoods} 
While the likelihood may be invariant to the scale of $A$ and $W$, the
Hessian term in equation~\ref{eqn:2ndorder} is not---intuitively, the
Hessian term gives local curvature information about the posterior and
encourages placing components in regions of high probability volume.
For different scales of $A$ and $W$, the isotropic variance
$\sigma^2_m$ may result in different relative volumes covered.  

The RRT nodes have much little scale flexibility as the entries of the
$\text{SVD}$ factors are fixed and for every change of basis matrix
$Q$ in the oblique manifold, the unit norm columns of the $Q$ matrix
determine a scaling as well.  Fortunately, we can easily rescale the
RRT components to cover an approximate volume.\footnote{The scale of
  $A$ and $W$ affects the prior term as well. Empirically, we observed
  that this rescaling affected the Hessian term more than the prior
  term, although this may not always be the case.}  We use the
analytical Hessian approximation for the squared error of the
factorization given in \cite{lin2007projected}. Given a component
$(A_m,W_m)$, under the Gaussian Likelihood with variance in the data
$X$ given by $\sigma^2_X$, the Hessian term is given by
$$
\text{Tr}(\bm{H}_m) = \frac{-1}{\sigma_X^2}\left(N\times\text{Tr}(A^TA) + D\times\text{Tr}(WW^T)\right)
$$ This term depends on the $l_2\text{-norm}$ of the columns of $A$
and the rows of $W$.  Let $S$ be the diagonal matrix that scales the
basis matrix to unit column norm so that
$\text{Tr}\left({(AS)^T(AS)}\right) = K$. We restrict our optimization
to the components where elements of the basis matrix have the same
squared column sum (so that column scales are consistent with each
other). Each component of this form looks like $(\beta A_mS,
\frac{1}{\beta}S^{-1}W_m)$. The Hessian term depends on $\beta$ in the
following way:
$$
\text{Tr}(\bm{H}_m)(\beta) = \frac{-1}{\sigma_X^2}\left(\beta NK + \frac{D}{\beta}\text{Tr}((S^{-1}W)(S^{-1}W)^T)\right)
$$ 
We use second order Newton Conjugate Gradient descent to perform an
optimization to find $\beta > 0$ that maximizes the Hessian term.

\section{Experimental Approach}

\subsection{Problem Formulation: Bayesian NMF} 
In our experiments, we consider exponential priors for the basis and
weights matrices: $p(A) = \prod_{d,k} \text{Exp}(A_{d,k};
\lambda_{A_{d,k}})$ and $p(W) = \prod_{k,n} \text{Exp}(W_{k,n};
\lambda_{W_{k,n}})$.  We consider two different likelihood models: 

\paragraph{Gaussian} The first is a Gaussian model in which Gaussian noise
is added to each dimension IID:
\begin{equation*} 
p_\mathcal{N}(X | A, W) =\prod_{d,n} \mathcal{N}(X_{d,n}, (AW)_{d,n}, \sigma_X^2)
\end{equation*} 
As derived in \cite{schmidt2009bayesian}, the combination of
exponential priors and Gaussian likelihoods results in a relatively
straightforward Gibbs sampling update.

\paragraph{Uniform} The light tails of the Gaussian distribution imply
that the samples far from the mode are quickly discounted.  In many
applications, a Uniform noise model may be more appropriate,
especially if the model is misspecified and we desire factorizations
that ``roughly'' model the data (that is, the difference between a
perfect factorization and one with some noise may not matter):
\begin{equation*}
p_\mathcal{U}(X | A, W) = \prod_{d,n} \mathcal{U}_{(-\epsilon, \epsilon)}(X_{d,n} - (AW)_{d,n})
\end{equation*}

\subsection{Baselines}
We compare to the following baselines: 
\begin{enumerate}
\item Standard Nonparametric Variational Inference (NVI) as described
  in \citet{gershman2012nonparametric} with $M=4$ and $M=10$
  components.
\item HMC + ONVI (HMC): We run Hamiltonian Monte Carlo with adaptive
  step size \citep{Neal} for 10,000 samples; each sample is proposed
  to the ONVI.
\item Gibbs + ONVI (Gibbs): For the Gaussian noise model only, we run
  the Gibbs sampler of \citet{schmidt2009bayesian} for 10,000 samples
  and propose each sample to the ONVI.  (There is no conjugate Gibbs
  sampler for the Uniform likelihood.) 
\end{enumerate}
The NVI components are initialized with a random solutions. In the
Uniform likelihood case, we initialize to a single feasible solution
as there is no gradient information otherwise.  The Gibbs and HMC
chains are initialized via first running Lin's algorithm
\citep{lin2007projected} on a random starting point to avoid the need
for burn-in period. We repeat each experiment 10 times and report the
mean as well as 25\textsuperscript{th} and 75\textsuperscript{th}
percentile of our obtained statistics.

\subsection{Implementation Details} 
\label{sec:imp_details}

\paragraph{ONVI Implementation}
We use second order Newton Conjugate Gradient descent to perform the
optimization. Gradients and Hessian-vector products of our objective
are computed using Autograd \citep{maclaurinautograd}.  

For the Gaussian likelihood, we use the second order approximation for
the likelihood term (equation~\ref{eqn:2ndorder}) to optimize both the
variance $\sigma^2_m$ of the new component as well as its weight
$w_m$.  Following optimization, if the new component improves the ELBO
by less than the same absolute change stopping criteria of $10e-4$
from \citet{gershman2012nonparametric}, we reject that component.
Similarly, if removing the previous components decreases the ELBO by
less than the absolute change criteria, we remove them.

The Uniform likelihood is flat within its feasible region and the
exponential prior also has no second order information.  Thus, we can
no longer rely on the Hessian term in equation~\ref{eqn:2ndorder} to
control the variance $\sigma^2_m$.  Instead, we fix the variance
$\sigma^2_m$ to the variance found by running standard variational
inference (via NVI with M=1) and optimize over the weights $w_m$.  We
also use a smarter stopping criteria corresponding to the increase in
entropy by a prospective component perturbed by $\sigma_m$ in every
dimension from an existing component.

\paragraph{RRT Implementation}
The framework of the RRT is very flexible and we customize it
depending on the likelihood model.  Below we describe the specifics of
our `Base Nodes', `Temporary Nodes' and feasibility criteria.

In Gaussian likelihood model the quality term dominates in the ELBO,
so we design the RRT to look for better quality components than
already present in its set of nodes. We initialize the RRT with $50$
solutions of Lin's algorithm. We also pick one of these and optimize
it under the ELBO approximation of equation~\ref{eqn:2ndorder}.  We
set the maximum number of `Temporary Nodes' to 100 and allow the RRT
to expand until it reaches that maximum. For subsequent feasible
components the RRT finds, we only replace them with the lowest quality
node in the tree. We initially set the feasibility threshold based on
the poorest quality node in the tree. Once the maximum number of
`Temporary Nodes' is reached, we increase the threshold of the
feasible region to be equal to that of the highest quality component.
There are no fixed `Base Nodes.'

In the Uniform likelihood model, the entropy term dominates in the
ELBO, so we design the RRT to look for components more diverse than
those in its existing set of nodes. We fix ten `Base Nodes' from
random restarts of Lin's algorithm and use these to repeatedly grow
the RRT.  For feasibility, we set the minimum angle condition to 0.01
degrees initially. We allow for 90 `Temporary Nodes' for the RRT's
expansion. Once the maximum number of nodes is reached, we re-start
the RRT with the `Base Nodes' and increase the minimum angle criteria
by 0.5. For any component that gets added to the ONVI, we add the
corresponding node to our set of `Base Nodes' as well.

We set the minimum step-size of the RRT in the oblique manifold to be $s_0 = 0.01$. This step-size grows by $10\%$ (for a maximum of 50 steps) during the expansion of the RRT as long as we are in a feasible region. In both noise models, we terminate the RRT+ONVI algorithm once the ONVI
has processed 5000 components or if the RRT fails to find a new
feasible node after 10,000 attempts at expansion.

\section{Results}

\subsection{Demonstration on Synthetic Data}
We embed a toy data matrix of rank $3$ known to have two exact NMFs
\citep{Laurberg} into $\mathbb{R}^{500 \times 500}$ and add Uniform
noise. This toy data set has an interesting property that only
slightly changing the data changes the nature of solutions from two to
infinite.  Thus, we expect that adding the noise will result in
multiple different solutions to exist within a certain threshold.  We
make the job of the RRT more difficult by only initializing it at one
(known) analytical solution.

\begin{figure}[tbh] 
\centering
\begin{minipage}{.45\textwidth}
  \centering
  \includegraphics[width=.9\linewidth]{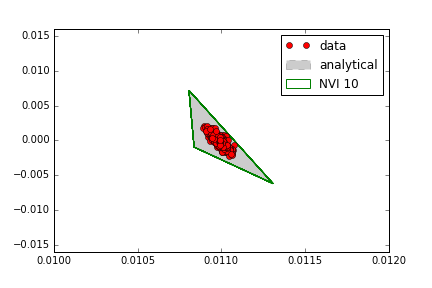}
  \captionof{figure}{A 2-D projection of Laurberg Data and   NVI components. The different components are  not distinguishable.}
  \label{fig:Laurberg_NVI10}
\end{minipage}%
\hfill
\begin{minipage}{.45\textwidth}
  \centering
  \includegraphics[width=.9\linewidth]{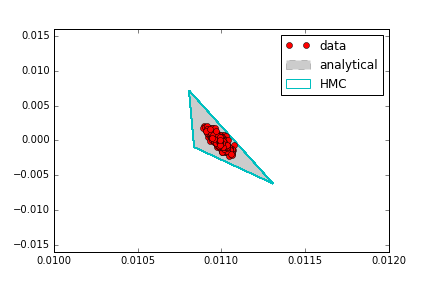}
  \captionof{figure}{A 2-D projection of Laurberg Data and the components accepted in the HMC + ONVI algorithm. These components are all coming from the same mode, which overlaps with the analytical solution (initialization).}
  \label{fig:Laurberg_HMC}
\end{minipage}

\begin{minipage}{.45\textwidth}
  \centering
  \includegraphics[width=.9\linewidth]{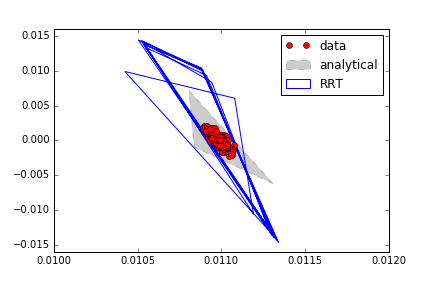}
  \captionof{figure}{A 2-D projection of Laurberg Data and the components accepted in the RRT + ONVI  algorithm. These components are distinct from the analytical solution (initialization). These  diverse components continue to explain the data under the Uniform noise model as the data lies in  the convex hull of these RRT-based factorizations.}
  \label{fig:Laurberg_RRT}
\end{minipage}%
\hfill
\begin{minipage}{.45\textwidth}
  \centering
  \includegraphics[width=.9\linewidth]{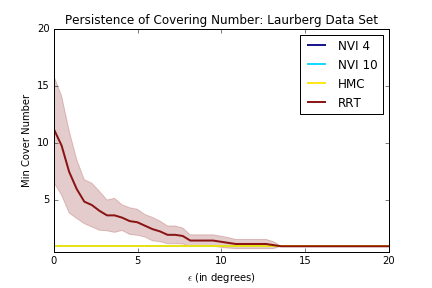}
  \captionof{figure}{Laurberg Data: A persistence plot of samples from the variational distribution obtained from the different algorithms. The RRT + ONVI is the only algorithm for which minimum cover numbers are larger than one for angles greater than 0.01. The other algorithms are unable to escape the modes that they were initialized in.}
  \label{fig:laurberg_cover}
\end{minipage}
\end{figure}

Figures~\ref{fig:Laurberg_NVI10},~\ref{fig:Laurberg_HMC},
~\ref{fig:Laurberg_RRT} show a two dimensional projection of the
rank-3 factorizations found by NVI (M = 10), ONVI+HMC, and ONVI+RRT.
Even when there are many factorizations, both NVI and HMC have very
limited exploration in comparison to the RRT.  The persistence plots
in figure~\ref{fig:laurberg_cover} echo this property: HMC and NVI
lines fall to 1 at an angle less than 0.01 while the RRT persists for
much longer.

\subsection{Application to Real Data Sets}
We apply our approach to several datasets commonly used for NMF \footnote{In the Reuters, and 20-Newsgroups data sets, we only took documents from the top 4 categories.}:
20-Newsgroups \citep{TwentyNewsgroups} (D=813, N=2034,
categories= R = 4); Reuters articles \citep{Reuters}(D = 1540, N = 2362, categories = R = 4); BBC articles
\citep{greene2006practical}(D = 6045, N = 1162, categories = R = 4); AML/ALL cancer cell data \footnote{The AML/ALL dataset actually has two categories. We chose the factorization rank to be 3 because it reveals useful sub-structure in the data \citep{brunet2004metagenes}}
\citep{vzitnik2012nimfa}(D = 5000, N = 38, R = 3) ; Olivetti Faces \citep{FacesData}(D = 4096, N = 400, R = 4); Hubble
telescope hyper-spectral image \footnote{Pixels in Hubble dataset were down-sampled from $N = 16384$)} \citep{HubbleData} (D=100, N=4096, R = 8).

\paragraph{RRT+ONVI give comparable or better ELBO values.}
Tables~\ref{table:elbo_small} and~\ref{table:elbo_large} show the ELBO
values for our approach compared to the baselines with the noise
$\sigma^2_X$ set to the empirical noise (found by first fitting a
solution via Lin's algorithm and measuring the squared error, before
any additional experiments) and 10x this empirical noise.  The latter
case was designed to simulate a situation where we might expect to see
greater variation in solutions.

In both cases, the RRT+ONVI has the highest ELBO across all data sets.
However, due to the light tails of the Gaussian distribution, the
expectation term (equation~\ref{eqn:2ndorder}) dominated the
(coverage-encouraging) entropy term (equation~\ref{eqn:ent}); all the
ONVI variants consistently retained only one highest quality
component.

Table~\ref{table:elbo_uniform} shows the results for the Uniform
likelihood (similar to before, the noise $\epsilon$ was set based on
the absolute deviation from fitting a solution via Lin's algorithm).
Here, the differences between solutions are smaller because they come
primarily from differences in the entropy term; thus we show the
deviation from the mean ELBO for each data set.  As before, the
RRT+ONVI finds the variational distribution with the highest ELBO.
The HMC+ONVI fails to move significantly from its starting point and
thus only retains one component; in contrast the RRT+ONVI retains
between 2 (AML) and 118 (BBC) components depending on the data set.

\begin{figure}[H] 
\centering
\begin{minipage}{.45\textwidth}
  \centering
  \includegraphics[width=.95\linewidth]{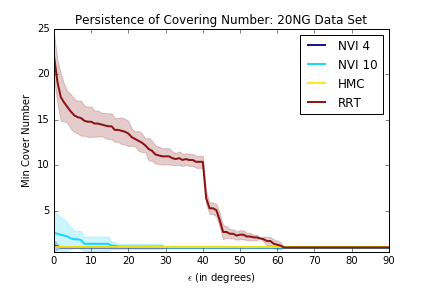}
  \captionof{figure}{20-Newsgroups data: Persistence plot of the components shows that RRT + ONVI components are more spread out than baselines.}
  \label{fig:20 Newsgroupspplot}
\end{minipage}%
\hfill
\begin{minipage}{.45\textwidth}
  \centering
  \includegraphics[width=.95\linewidth]{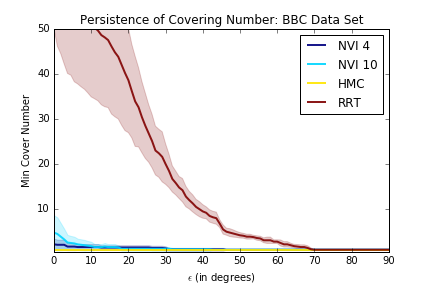}
  \captionof{figure}{BBC data: Persistence plot of the components shows that RRT + ONVI components are more spread out than baselines.}
  \label{fig:BBCpplot}
\end{minipage}

\begin{minipage}{.45\textwidth}
  \centering
  \includegraphics[width=.95\linewidth]{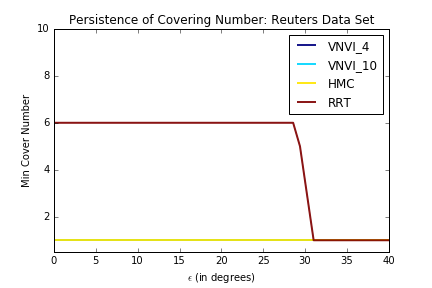}
  \captionof{figure}{Reuters data: Persistence plot of the components shows that RRT + ONVI components are more spread out than baselines.}
  \label{fig:Reuterspplot}
\end{minipage}%
\hfill
\begin{minipage}{.45\textwidth}
  \centering
  \includegraphics[width=.95\linewidth]{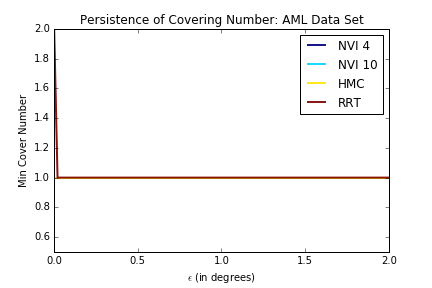}
  \captionof{figure}{AML/ALL data: The RRT + ONVI components are more spread out than baselines but it is a much smaller spread in terms of angle. For angles greater than 0.1 the minimum covering number is one for all algorithms.}
  \label{fig:AMLpplot}
\end{minipage}

\begin{minipage}{.45\textwidth}
  \centering \includegraphics[width=.95\linewidth]{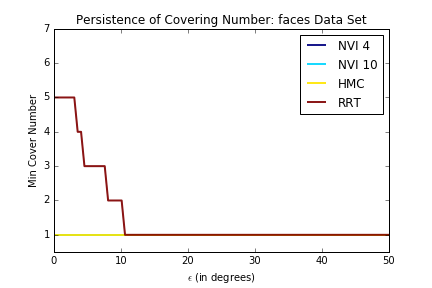}
  \captionof{figure}{Faces data: The RRT + ONVI components are spread
    out more than baselines.}
  \label{fig:Facespplot}
\end{minipage}%
\hfill
\begin{minipage}{.45\textwidth}
  \centering
  \includegraphics[width=.95\linewidth]{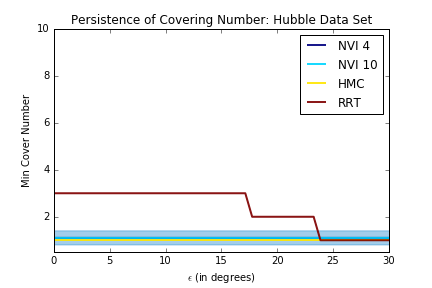}
  \captionof{figure}{Hubble data: The RRT + ONVI components show the largest posterior coverage.}
  \label{fig:Hubblesmallpplot}
\end{minipage}
\end{figure}

\paragraph{RRT+ONVI has more posterior coverage}
The evaluations above showed that our RRT+ONVI approach achieves
comparable or better ELBO values on real datasets.  Now we return to
our goal: posterior coverage.  Figures~\ref{fig:20 Newsgroupspplot},
~\ref{fig:BBCpplot}, ~\ref{fig:Reuterspplot}, ~\ref{fig:AMLpplot},
~\ref{fig:Facespplot}, and~\ref{fig:Hubblesmallpplot} plot the minimum
covering number of the components of the variational distribution
under the WAD similarity measure. While there is a range of variation
across the data sets in terms of angle, the RRT components are most
distinctly placed.

\begin{table}[hb]
\begin{minipage}{\textwidth}
\caption{Empirical Gaussian Noise Setting - We show values of the mean ELBO as well as the 25\textsuperscript{th} and 75\textsuperscript{th} percentiles.  Across the datasets, the RRT + ONVI procedure finds the best variational fit to the true posterior.}
\label{table:elbo_small}
\renewcommand{\arraystretch}{0.7}
\begin{center}
\begin{tabular}{|c|c|c|c|c|c|}
\hline
&\parbox{1.7cm}{\centering NVI-4}&\parbox{1.7cm}{\centering NVI-10}&\parbox{1.7cm}{\centering Gibbs}&\parbox{1.7cm}{\centering HMC}&\parbox{1.7cm}{\centering RRT}\\ 
\hline
\multirow{3}{*}{{20 Newsgroups}} &3.115e+06&3.116e+06&3.145e+06&3.151e+06&3.191e+06\\
 &\tiny3.115e+06, &\tiny3.115e+06, &\tiny3.144e+06, &\tiny3.150e+06, &\tiny3.190e+06, \\
 &\tiny3.116e+06&\tiny3.117e+06&\tiny3.145e+06&\tiny3.152e+06&\tiny3.191e+06\\
\hline
\multirow{3}{*}{{BBC}} & 2.030e+07&2.030e+07&2.058e+07&2.051e+07&2.061e+07\\
 &\tiny2.030e+07, &\tiny2.030e+07, &\tiny2.057e+07, &\tiny2.049e+07, &\tiny2.061e+07, \\
 &\tiny2.030e+07&\tiny2.030e+07&\tiny2.058e+07&\tiny2.052e+07&\tiny2.061e+07\\
\hline
\multirow{3}{*}{{Reuters}} &8.266e+06&8.266e+06&8.239e+06&8.232e+06&8.276e+06\\
 &\tiny8.264e+06, &\tiny8.265e+06, &\tiny8.235e+06, &\tiny8.228e+06, &\tiny8.264e+06, \\
 &\tiny8.267e+06&\tiny8.267e+06&\tiny8.243e+06&\tiny8.237e+06&\tiny8.290e+06\\
\hline
\multirow{3}{*}{{AML}} &-1.536e+06&-1.536e+06&-1.571e+06&-1.564e+06&-1.510e+06\\
 &\tiny-1.536e+06, &\tiny-1.536e+06, &\tiny-1.571e+06, &\tiny-1.565e+06, &\tiny-1.510e+06, \\
 &\tiny-1.536e+06&\tiny-1.536e+06&\tiny-1.570e+06&\tiny-1.563e+06&\tiny-1.510e+06\\
\hline
\multirow{3}{*}{{Faces}} &1.292e+06&1.293e+06&1.284e+06&1.288e+06&1.314e+06\\
 &\tiny1.290e+06, &\tiny1.292e+06, &\tiny1.281e+06, &\tiny1.283e+06, &\tiny1.314e+06, \\
 &\tiny1.295e+06&\tiny1.295e+06&\tiny1.288e+06&\tiny1.294e+06&\tiny1.314e+06\\
\hline
\multirow{3}{*}{{Hubble}} &-2.593e+07&-8.079e+06&-8.022e+06&-6.179e+07&1.975e+06\\
 &\tiny-3.566e+07, &\tiny-1.199e+07, &\tiny-1.554e+06, &\tiny-8.485e+07, &\tiny1.959e+06, \\
 &\tiny1.988e+06&\tiny1.815e+06&\tiny4.441e+05&\tiny-1.562e+07&\tiny1.990e+06\\
\hline
\end{tabular}
\end{center}
\end{minipage}
\begin{minipage}{\textwidth}
\caption{Larger Gaussian Noise Setting - We show values of the mean
  ELBO as well as the 25\textsuperscript{th} and
  75\textsuperscript{th} percentiles.  Across the datasets, the RRT +
  ONVI procedure finds the best variational fit to the true
  posterior.}
\label{table:elbo_large}
\renewcommand{\arraystretch}{0.7}
\begin{center}
\begin{tabular}{|c|c|c|c|c|c|}
\hline
&\parbox{1.7cm}{\centering NVI-4}&\parbox{1.7cm}{\centering NVI-10}&\parbox{1.7cm}{\centering Gibbs}&\parbox{1.7cm}{\centering HMC}&\parbox{1.7cm}{\centering RRT}\\ 
\hline
\multirow{3}{*}{{20 Newsgroups}}&2.298e+05&2.298e+05&1.968e+05&2.050e+05&2.326e+05\\
 &\tiny2.297e+05, &\tiny2.298e+05, &\tiny1.967e+05, &\tiny2.049e+05, &\tiny2.326e+05, \\
 &\tiny2.298e+05&\tiny2.299e+05&\tiny1.970e+05&\tiny2.050e+05&\tiny2.327e+05\\
\hline
\multirow{3}{*}{{BBC}} &7.709e+06&7.712e+06&7.782e+06&7.760e+06&7.844e+06\\
 &\tiny7.693e+06, &\tiny7.698e+06, &\tiny7.781e+06, &\tiny7.760e+06, &\tiny7.844e+06, \\
 &\tiny7.726e+06&\tiny7.723e+06&\tiny7.782e+06&\tiny7.761e+06&\tiny7.844e+06\\
\hline
\multirow{3}{*}{{Reuters}} &1.767e+06&1.767e+06&1.723e+06&1.730e+06&1.768e+06\\
 &\tiny1.767e+06, &\tiny1.767e+06, &\tiny1.723e+06, &\tiny1.730e+06, &\tiny1.768e+06, \\
 &\tiny1.767e+06&\tiny1.767e+06&\tiny1.724e+06&\tiny1.730e+06&\tiny1.768e+06\\
\hline
\multirow{3}{*}{{AML}} &-1.842e+06&-1.842e+06&-1.858e+06&-1.873e+06&-1.818e+06\\
 &\tiny-1.843e+06, &\tiny-1.842e+06, &\tiny-1.859e+06, &\tiny-1.874e+06, &\tiny-1.818e+06, \\
 &\tiny-1.842e+06&\tiny-1.842e+06&\tiny-1.858e+06&\tiny-1.872e+06&\tiny-1.818e+06\\
\hline
\multirow{3}{*}{{Faces}} &-1.644e+06&-1.643e+06&-1.647e+06&-1.643e+06&-1.631e+06\\
 &\tiny-1.644e+06, &\tiny-1.643e+06, &\tiny-1.648e+06, &\tiny-1.643e+06, &\tiny-1.631e+06, \\
 &\tiny-1.644e+06&\tiny-1.643e+06&\tiny-1.647e+06&\tiny-1.642e+06&\tiny-1.631e+06\\
\hline
\multirow{3}{*}{{Hubble}} &7.551e+05&9.657e+05&1.126e+06&6.692e+05&1.208e+06\\
 &\tiny8.755e+05, &\tiny9.347e+05, &\tiny1.119e+06, &\tiny8.575e+05, &\tiny1.207e+06, \\
 &\tiny1.122e+06&\tiny1.089e+06&\tiny1.136e+06&\tiny9.407e+05&\tiny1.209e+06\\
\hline
\end{tabular}
\end{center}
\end{minipage}
\end{table}

\newpage

\begin{table}[ht]
\caption{Uniform Likelihood: This table shows ELBO deviation from dataset mean and number of RRT components accepted into the ONVI. In all cases, the HMC + ONVI led to unimodal variational distributions. We see that the range of components found are data-set dependent. We provide the mean, first quartile (25) and third quartile (75).}
\label{table:elbo_uniform}
\renewcommand{\arraystretch}{0.7}
\begin{center}
\begin{tabular}{|c|c|c|c|c|c|}
\hline
&\parbox{1.7cm}{\centering NVI 4}&\parbox{1.7cm}{\centering NVI 10}&\parbox{1.7cm}{\centering HMC}&\parbox{1.7cm}{\centering RRT}&\parbox{1.9cm}{\centering Components (RRT)}\\ 
\hline
\multirow{3}{*}{{20 Newsgroups}} &-3.080e-01&-1.228e+06&-1.499e+00&1.693e+00&14.8\\
 &\tiny-1.708e+00, &\tiny-1.228e+06, &\tiny-1.708e+00, &\tiny1.823e+00, &\tiny8, \\
 &\tiny3.500e-01&\tiny-1.228e+06&\tiny-1.360e+00&\tiny1.883e+00&\tiny21\\
\hline
\multirow{3}{*}{{BBC}} &-2.780e-01&-3.481e+06&-3.109e+00&3.040e+00&57\\
 &\tiny-1.210e+00, &\tiny-3.481e+06, &\tiny-3.117e+00, &\tiny3.035e+00, &\tiny29.2, \\
 &\tiny9.375e-01&\tiny-3.481e+06&\tiny-3.103e+00&\tiny3.050e+00&\tiny77.2\\
\hline
\multirow{3}{*}{{Reuters}} &-1.840e-01&-2.079e+06&-7.622e-01&1.030e+00&6\\
 &\tiny-7.675e-01, &\tiny-2.079e+06, &\tiny-7.700e-01, &\tiny1.030e+00, &\tiny6 \\
 &\tiny6.825e-01&\tiny-2.079e+06&\tiny-7.600e-01&\tiny1.030e+00&\tiny6\\
\hline
\multirow{3}{*}{{AML}} &6.610e-01&-2.216e+06&-2.749e+00&1.430e+00&2\\
 &\tiny6.225e-01, &\tiny-2.216e+06, &\tiny-6.585e+00, &\tiny1.430e+00, &\tiny2, \\
 &\tiny7.275e-01&\tiny-2.216e+06&\tiny7.025e-01&\tiny1.430e+00&\tiny2\\
\hline
\multirow{3}{*}{{Faces}} &-4.580e-01&-8.713e+05&-4.580e-01&1.370e+00&5\\
 &\tiny-5.825e-01, &\tiny-8.713e+05, &\tiny-5.825e-01, &\tiny1.370e+00, &\tiny5, \\
 &\tiny-2.800e-01&\tiny-8.713e+05&\tiny-2.800e-01&\tiny1.370e+00&\tiny5\\
 
\hline
\multirow{3}{*}{{Hubble}} &-5.389e-01&-1.563e+05&-5.389e-01&1.620e+00&3\\
 &\tiny-1.170e+00, &\tiny-1.563e+05, &\tiny-1.170e+00, &\tiny1.620e+00, &\tiny3, \\
 &\tiny-3.800e-01&\tiny-1.563e+05&\tiny-3.800e-01&\tiny1.620e+00&\tiny3\\
\hline
\end{tabular}
\end{center}
\end{table}

In particular, the persistence plots that for datasets such as
20-Newsgroups and BBC have a lot of variation in the number of modes
found and some variation in how the persistence curve drops to 1 as
angle gets larger. This variation is indicative of the random nature
of the RRT across different repetitions of the experiment. In the
remaining datasets we see much less variation. It appears that the
ONVI has consistently accepted components from the random restarts of
Lin's algorithm (in the initialization of the RRT). In the case of AML
and Faces datasets, the RRT finds multiple components for the ONVI to
consider but the ONVI rejects them based on the entropy-based ELBO improvement
criteria. In the Reuters and Hubble datasets, the RRT fails to expand
from the initial base nodes.

\section{Discussion} 
Our rapid posterior exploration approach involved making many choices,
ranging from the approximating variational family to the exploration
properties of the RRT.  First and foremost, we observed that the the
choice of the likelihood model plays a crucial role in determining the
shape of the posterior---obvious, perhaps, but not necessarily
observable if one's algorithms limit the posterior approximation to
small regions.  The noise parameters can also have large effects.  For
consistency, in our experiments we set $\epsilon$ in the Uniform
likelihood as the largest absolute deviation between the data and NMF
approximation based on ten random restarts of Lin's algorithm.
However, in other experiments, we found increasing $\epsilon$---to
$\epsilon_{\text{new}} = 2\epsilon$ for Faces,
$\epsilon_{\text{new}}=8\epsilon$ for Hubble---resulted in many more
components in the posterior contructed by ONVI.  It may be more
appropriate to specify $\epsilon_{i,j}$ for each entry $i,j$ in the
data $X$.

The isotropic variance of our variational family is restrictive. For high-density regions of the Bayesian NMF posterior, we can expect a perturbation of the basis matrix to be correlated to an appropriate change in the weights matrix. A mixture of Gaussians with more general covariances might therefore be
most appropriate.  For such a model, \citet{huber2008entropy} describe
a generalization of the lower bound of the entropy formula that we
used in equation~\ref{eqn:ent}. The question of determining the
entries of the covariance matrices would need some insight. Simply
trying to optimize for the entries in the covariance matrix would make
the problem high-dimensional again as each step in the ONVI would
involve solving for a covariance matrix of a $R(D + N)-$dimensional
vector.  

\newpage 

The ONVI serves to  accept/reject candidate components and to adjust accepted components within the variational posterior in an optimal manner. Before introducing candidate components to the ONVI, one might also imagine using a diversification strategy such
as hard-core processes or determinental point processes
\cite{kulesza2012determinantal} to cull the set of candidate nodes. Such a step could help eliminate redundant candidate nodes and reduce the number of candidate nodes presented to the ONVI.

Lastly, there were many parameter choices made in the initialization and expansion of the RRT. Changing the RRT parameters can make a difference in the posterior constructed by the ONVI. For example, the RRT fails to expand in the Reuters dataset (under the Uniform likelihood) using our default parameters, but we noticed that increasing the initial step-size from $s_0 = 0.01$ to $s_0 = 0.5$ allows the RRT to expand and contribute new components to the posterior constructed by the ONVI. An interesting future direction is to adapt the RRT to specific noise-models and datasets. To allow for a larger number of temporary and base nodes (to aid the RRT expansion) we suggest using more efficient data structures for finding the nearest point in the tree \citep{atramentov2002efficient}. For exploring posterior regions that are not conducive to RRT expansion, alternative exploratory techniques such as probabilistic road maps \citep{kavraki1996probabilistic} can be employed.

\section{Related Work}
Outside of the empirical work of \cite{Greene} and
\cite{roberts2016navigating}, we are not aware of work that has been
done to to explicitly consider multiple modes (or connected posterior
regions) in NMF.  In the Bayesian setting, there have been several
general approaches to encourage samplers to explore and mix between
modes. \cite{Neal} discusses a specific approach to tempering in HMC;
\cite{green2001delayed} discusses an MH approach that delays
rejections to allow a far-reaching proposal to find an alternate mode.
\cite{sminchisescu2007generalized} describe approaches for hopping
between modes if the locations of the modes are known to the sampler
in advance. Both simulated \citep{marinari1992simulated} and parallel
\citep{hansmann1997parallel} tempering methods are popular for
encouraging samplers to move to new regions of high posterior
density. Wang-Landau sampling \citep{schulz2003avoiding} is a popular
MCMC sampling method in statistical physics that involves a random
walk in the energy space rather than sampling at a fixed temperature.

Within variational methods, mean-field approaches
\citep{beal2003variational, jordan1999introduction} are popular but
limited to approximating posteriors with unimodal distributions. The use of mixture distributions in mean-field approaches \citep{lawrence1998approximating,jaakkola1998improving,bouchard2009split} adds flexibility to the posterior but the parameter updates are model dependent and may be difficult to derive. We choose a Gaussian mixture with isotropic variance for our variational family. This is a special case of the mixture mean-field approach for which the inference algorithm provided by \citep{gershman2012nonparametric} can be used for a more general class of models as only knowing the joint likelihood and its gradient is required. Flow-based methods of creating flexible posterior families (such as Non-linear Independent Components Estimation  \citep{dinh2014nice} and Hamiltonian variational approximation \citep{salimans2015markov} use a series of functions that transform the parameter space. The normalizing flows framework unifies and presents a generalization of these flow-based variational methods for flexible posteriors  \citep{rezende2015variational}

More broadly, the problem of locating multiple modes or a frontier of solutions has been studied
in the optimization literature, though not for the particular case of
NMF.  Many of these approaches are particle or swarm approaches, in
which multiple solutions are initialized, adjusted, and killed according
to some evolution and fitness function
(e.g. \cite{dorsey1995genetic,brits2007locating}.  In some cases,
explicit rules are made for finding solutions that are in alternate
modes \citep{wales1997global}.  Homotopy methods are used to explore the Pareto surface of nonlinear functions \citep{das1998normal}.

In some work, NMF is specifically used for clustering data. \citet{Greene} and \citet{huang2011enhanced} use ensemble NMF techniques to understand the space of clusterings obtained through distinct NMF solutions (obtained from random restarts of NMF alogirthms). More work exists for finding multiple alternate clusterings, which can
be viewed as a subset of matrix factorizations in which each
observation is associated with only one latent feature.
\cite{niu2010multiple} iteratively find multiple clustering views by
taking advantage of relationships between the spectral clustering
objective and the Hilbert-Schmidt Independence Criterion.
\cite{qi2009principled} apply constrained optimization that defines
trade-offs between alternativeness and clustering quality, also
separate into finding a single alternative and multiple alternatives,
while \cite{gondek2007non} use a mutual information-based criterion to
subtract out the information from existing alternatives to propose new
ones.  Finally, \cite{grimmer2011general} simply run a large number of
different clustering algorithms and display the alternatives.

To our knowledge, our application of
explicit exploration algorithms for posterior coverage is novel.

\section{Conclusion}
In this work, we leveraged some key geometric insights about NMF to
first create a smaller search space, applied RRTs to explore this
space, and incorporated these nodes into a flexible, more complete
posterior via the nonparametric variational inference framework.
Importantly, only the design of the RRT using the oblique manifold is
specific to NMF: we expect that this notion of casting posterior
inference as an explicit exploration problem will be fruitful for many
other models as well.

\subsubsection*{Acknowledgements}
We thank Andrew Miller, Wei Wei Pan, and Michael Hughes for insightful
discussions.  This work was supported by Defense Advanced Research
Projects Agency (DARPA) grant number W911NF-16-1-0561.

\bibliographystyle{unsrtnat}
\bibliography{ONVI_NMF}
\end{document}